\documentclass{article}
\usepackage{ijcai21}

\usepackage{times}
\usepackage{soul}
\usepackage{url}
\usepackage[hidelinks]{hyperref}
\usepackage[utf8]{inputenc}
\usepackage[small]{caption}
\usepackage{graphicx}
\graphicspath{ {./images/} }
\usepackage{amsmath}
\usepackage{amsfonts}
\usepackage{amsthm}
\usepackage{booktabs}
\usepackage{algorithm}
\usepackage{algorithmic}
\usepackage{subcaption}

\usepackage{xcolor}

\makeatletter
\newcommand*\titleheader[1]{\gdef\@titleheader{#1}}
\AtBeginDocument{%
\let\st@red@title\@title
\def\@title{%
\bgroup\normalfont\large\centering\@titleheader\par\egroup
\vskip1.5em\st@red@title}
}
\makeatother

\title{End-To-End Anomaly Detection for Identifying Malicious Cyber Behavior through NLP-Based Log Embeddings}

\titleheader{\emph{IJCAI-21 1\textsuperscript{st} International Workshop on Adaptive Cyber Defense}}

\author{
Andrew Golczynski
\And
John A. Emanuello
\affiliations
Laboratory for Advanced Cybersecurity Research\\
National Security Agency\\
}

\begin{document}

\maketitle

\begin{abstract}
 Rule-based IDS (intrusion detection systems) are being replaced by more robust neural IDS, which demonstrate great potential in the field of Cybersecurity. However, these ML approaches continue to rely on ad-hoc feature engineering techniques, which lack the capacity to vectorize inputs in ways that are fully relevant to the discovery of anomalous cyber activity. We propose a deep end-to-end framework with NLP-inspired components for identifying potentially malicious behaviors on enterprise computer networks. We also demonstrate the efficacy of this technique on the recently released DARPA OpTC data set.
\end{abstract}
\section{Introduction}

\indent\indent Automated IDS (intrusion detection systems) have been of interest to researchers and cybersecurity professionals for several decades \cite{denning87}.  Earlier generations of IDS utilities leveraged manual rules and traditional machine learning techniques paired with ad-hoc feature engineering to reproduce expert capabilities given limited processing power and data availability.  However, these hard-coded features and signature-based rules failed to be generally robust (especially when new threats are introduced), and traditional machine learning models generally lacked the capacity to sufficiently discern anomalous behaviors in large enterprise computer networks \cite{verma18}.  Improved IDS implemented via neural-based machine learning has been long-expected, and may have the capacity to address these issues \cite{lunt90}, but such systems have only become feasible in the past several years due to recent improvements in hardware and data collection. Autoencoder-based anomaly detection has been applied in this space with some success, enabling the discovery of malicious activity under the assumption that such activity will differ markedly from the ordinary users of a network or other system \cite{hashemi2020}.

With such modern approaches coming to bear, we can now expect the development of a more-capable family of neural IDS.  However, there remain several issues in the design and implementation of these systems; in particular, autoencoder-based IDS typically still rely on the frail ad-hoc feature\footnote{We differentiate \emph{fields}, which are found in raw log data, and \emph{features}, which refer to the output of some sort of process and which give a meaningful representation of the raw data.}  engineering used by traditional IDS.  To truly bring these neural systems to fruition, a more automated and robust approach to record vectorization must be developed.

To that end, this paper details our work, which was inspired by natural language processing (NLP) techniques, to provide autonomous feature generation of network and host log metadata vectorization.  By treating logs' entry fields as tokens in a text corpus, we seek to remove the assumptions made by traditional feature engineering approaches, instead allowing our utility to \emph{learn} vector representations of the in our log files, leading to natural embeddings of records describing system activities in a manner \emph{relevant to the description and discovery of anomalous behavior in a network} (e.g. which IPs commonly commonly communicate with each other and which dlls are commonly loaded for particular applications).  Furthermore, we train these embeddings in-line with our anomaly detection utility, forming an end-to-end architecture with embeddings that are specifically trained to improve performance on our downstream anomaly detector, ensuring relevance and performance.

In this work, we will provide a brief overview of the DARPA Operationally Transparent Computing (OpTC) dataset leveraged in this paper, including discussing some of the relevant aspects of  data labeling and pre-processing/engineering of the data.  We will then discuss our architectural motivations and choices, along with implementation, training and testing details for our proof-of-concept utility.  The results of this work will then be presented and discussed, followed by concluding remarks.

\subsection{Background and Related Work}

\indent\indent A typical traditional approach to log-based intrusion detection is provided by Verma and Bridges \cite{verma18} -- the authors develop a metric space for describing Windows host logs, but their methodology relies on ad-hoc features and metrics rather than a learned/neural approach.  Other authors, such as Chen et al., have worked to create more automated feature engineering using NLP-based vectorization, but their intersection with NLP also relies on non-neural traditional approaches, namely term frequency inverse document frequency (TF-IDF) vectorizations of log files \cite{chen19}, which lack the ability to dynamically learn fit-for-purpose representations of data.  Di Mauro et al recently provided a survey of neural approaches to intrusion detection, showing the potential of neural techniques in this space, but none of the referenced works looks to modern word embeddings as a solution to our noted feature engineering problem \cite{mauro20}.

Statistical anomaly detection approaches, such as those in Heard \& Rubin-Delanchy \cite{heard16}, and Whitehouse et al. \cite{whitehouse16} do show promise, but do not appeal to neural network based approaches such as Nguyen et al. \cite{nguyen19}, which have a wider capacity to learn more complex representations of baseline cyber behavior.

Here, the idea is to train the neural network architecture, such as an autoencoder (AE). AEs are trained in an unsupervised fashion, wherein the target function to be learned is an approximate identity $f: \mathbb{R}^n \rightarrow \mathbb{R}^n$ on the input space\cite{goodfellow2016}. These networks are usually decomposed into non-linear encoder $ h: \mathbb{R}^n \rightarrow \mathbb{R}^m$ and decoder $g:  \mathbb{R}^m \rightarrow \mathbb{R}^n$ functions such that $m<<n$.
By construction, AEs learn salient information about the training data and reconstruction will fail on outliers or data drawn from distributions vastly different from that of the training set, making them ideal for anomaly detection (using ($L_2$) reconstruction error $\left\|x-f(x)\right\|_{2}$ as an anomaly score)\cite{cozzolino16,liao18}.

Our work borrows from the modern family of neural word embedding techniques, developed by researchers such as Bengio et al. \cite{bengio06} and Mikolov et al. \cite{mikolov13,mikolov132} in their development of the word2vec-style utilities and their related extensions to embed words in a vector space in which semantic relationships between words are encoded as geometric properties in the embedded space.  While their original intention was strictly rooted in text analysis, other works have shown that the word2vec style of embedding can be applied to other domains to produce coherent and useful embeddings -- for instance, see Perozzi et al. \cite{perozzi14}, wherein the authors leverage word2vec as a method for embedding graph representations, as well as Liu et al., wherein a similar vectorization approach (log2vec) is utilized to embed log entries as nodes of a graph \cite{liu19}.  An application of this approach can be found in Ring et al., which leverages word2vec-style embeddings to embed the space of IP addresses for network traffic analysis \cite{ring17}.  Similarly, Ramstr{\"o}m et al. and Wong et al. use a word2vec-inspired vectorization of netflow features as input for an AE-based anomaly detection scheme \cite{ramstrom19,wong21}. The latter work also demonstrated that the AEs were quite robust to the presence of data corresponding to cyber attacks in the training data, which is likely due to the AE's ability to ignore characteristics of rare events.

\section{Data}

\indent\indent The availability of quality benchmark datasets concerning malicious cyber activity is a perennial challenge for researchers in this space -- cyber activity is generally complex and multifaceted, meaning that any one dataset can only describe a small portion of malicious cyber behavior.  Additionally, labeling such data sets can be a challenge, as combing through millions of logs to determine which logs are actually associated with legitimate malicious activity is a daunting task at the best of times, leading to labeling that can be incomplete or incorrect.  Some datasets \cite{moustafa15} rely on synthetic data to provide greater variety and accuracy, but these may fail to capture the nuances of both normal user activity and actual attack patterns, and may overly-simplify downstream machine learning challenges.  

For this work, we relied on the Operationally Transparent Cyber Dataset created by Defense Advanced Research Projects Agency (DARPA) \cite{optc19}, which was designed to help address some of the above issues. These data consist of  network and host logging collected from a network of hundreds of Windows hosts over a one-week period, representing normal (benign) user activity.  Over a three-day period within this week, a set of red team actors also worked to perform various penetration tests against the network, seeking to infiltrate the network, ensure persistence, and carry out increasingly-complex attacks over time.  The data we sought to use consisted of the following:
\begin{enumerate}
	\item Combined network and host logs in the ECAR (Extended Cyber Analytic Repository) format, aggregated by host and time.
	\item An description of the attacks carried out by the red team actors, with timestamps, Windows process IDs, hostnames, and general notes.
\end{enumerate}

With over 17B events (from approximately 500 hosts) corresponding to detailed, security-oriented network and host logging, along with detailed descriptions of the red team activities on the network, the open source OpTC dataset provides: (1) a more realistic scenario (multi-staged redteam attacks) on which to test our cyber activity detection task; and (2) a means for others to validate our results and to compare our technique with theirs\footnote{See Anjum et. al for detailed analysis of the data and a discussion on its utility for ML-based cybersecurity research activities\cite{Anjum_2021}.}.

\subsection{Data Format}

\indent\indent The ECAR data format is an extension of Mitre's CAR format, which consists of nested JSON dictionaries, with each record primarily describing an object (e.g. a process, thread, registry key, etc.) and the action taken on that object (e.g. starting a process, modifying a registry value, etc.).  Additional object/action-specific properties are populated on the record to provide necessary context as well (e.g. a record pertaining to the editing of a registry value would have an optional property describing the new value being set) \cite{car20}.  The ECAR format simply adds additional object/action pairs that provide a greater level of detail than the original CAR \cite{optc19}.  Overall, the ECAR data contains 58 unique keys in its nested structure, with each record only populating a minority of those fields.

The OpTC ECAR dataset consists of roughly 10 terabytes of data uncompressed, with tens of millions of records per host across the collection period \cite{Anjum_2021}.  This provides fertile ground for our modeling approaches, while still having a relatively-constrained vocabulary designed for unambiguous logging rather than expressive natural language.

\subsection{Train/Test Split and Labeling}

\indent\indent Fundamentally, our goal is to train an anomaly detector on baseline activity that can then be applied to later activity to detect abnormal behavior.  To mirror this need, we used the days of benign activity as our training data set, and reserved the days containing both benign and red team activity for testing, with the goal of detecting the red team activity in our test set.

However, the records in the ECAR data are \emph{not} inherently labeled with benign/malicious tags.  To address this, we developed labels for the data using the description of the malicious activities provided by the red team actors on their active days.  For our purposes, an ECAR record is considered to be malicious if it is one of the malicious agent processes noted by the red team, or was spawned by a malicious process.  Additionally, the data was searched for any miscellaneous malicious network activity that was not directly associated with the noted processes -- while the majority of the records we labeled as "malicious" were discovered via the former standard, a small minority were caught through the latter (specifically, activity associated with ping sweeps performed on the network were problematic, given that the processes performing such actions were associated with system utilities, rather than being spawned directly by the malicious agents).

This labeling approach, it should be noted, casts a broad net -- all logs associated with the clearly-malicious activity are discovered by this approach, but generic activity that is incidentally associated with the malign activity is also caught.  While this produces somewhat noisy labels, our main concern is not to achieve 100 percent accurate classification on a record-by-record basis, but rather to catch the overarching attacks as described by activity across multiple logs with a high degree of precision -- primarily, we aim to provide the SOC admin strong evidence of behavior indicative of an attack, rather than providing them with all activities that may be associated with an attack.  Given this goal, the incidentally-malicious logs being labeled as malign is of minimal concern to us.

\section{Data Pre-processing}

\indent\indent For this work, we created a \emph{vocabulary}  consisting of terms that were tokenized and/or pre-processed versions of field values\footnote{In keeping with terminology associated dictionary data types such as JSON, we will use the terms \emph{key} (often inter-changeably with \emph{field}) and the precise  instances these fields may take on as in logs as \emph{values}.} from the ECAR records.  Our pre-processing philosophy was to have as light as touch as possible, which should allow us to avoid the pitfalls associated with the largely ad-hoc feature engineering approaches that are standard in IDS.  A proof of concept that only alters its input minimally should show that the text processing techniques are responsible for the bulk of the learning, rather than the guilty knowledge of the researchers being the driving factor behind our efficacy.  In total, our pre-processing consisted of the following steps:

\begin{enumerate}

	\item Dropping unnecessary features

		Of the fifty-eight keys in the ECAR file, only twenty-seven were kept for processing.  The remaining fields were determined to be insufficiently relevant to our chosen task/approach, and were dropped (see Appendix \ref{app:ECAR}. for the complete list of keys that we used.)

	\item Feature prefixing

		Some terms may have specific meanings depending on which feature with which they are associated.  For example, the number 443 has a specific meaning in the dest\textunderscore port field, but that meaning would not be preserved in other fields.  To ensure that these meanings are respected, values of select features were prefixed with their feature name (e.g. "443" would be mapped to the term "PORT\textunderscore 443" when it is found in the dest\textunderscore port field).

	\item Connection time bucketing
		
		Network activity records contain start and end times for the connection; to discretize these values, connection durations were calculated and bucketed into SMALL, MEDIUM, and LARGE buckets based on manual inspection of the distribution of connection durations.

	\item Path/file name extraction

		File paths and registry keys can contain machine specific sets of directories, even when considering a common item (e.g. "xyz.dll" might be found in separate directory locations on separate machines, even though the underlying file is the same).  To correct for this, such paths are reduced to only the file name.  Similarly, when parsing command line input, only the name of the executed program is maintained (i.e. the full path and any arguments are dropped).

	\item /24 CIDR subnet extraction

		IPv4 addresses were, in many cases, too sparse across the dataset to be well-represented, so we opted to remove the last octet from each address and only use the /24 subnet.  While this did reduce the resolution of the representation of our addresses, we gained additional robustness by aggregating rare terms into a smaller number of representations. 

	\item Emphemeral port aggregation

		Ports used for outgoing network traffic are typically arbitrarily chosen from the high end (greater than 49151) of the range of valid ports.  These choices carry no real meaning, and needlessly expand the vocabulary space, and so such ports were replaced with a generic "EPHEMERAL\textunderscore PORT" token.

	\item Removal of rare terms

		After performing the above tokenization, terms that appeared fewer than ten times (determined by examination of the distribution of term frequencies) over our training corpus were discarded and replaced with a generic "OBSCURE\textunderscore TERM" token.

	\item Conversion to vocabulary indices

		All remaining vocabulary tokens were indexed, and each token was replaced with its index.  Empty fields (e.g. file path fields are empty for network traffic events) were replaced with a "NULL\textunderscore TERM" token.

\end{enumerate}

Given the sheer volume and variability of our data, we additionally elected to focus proposed deep learning training regimen on each of four hosts (201-204) in the dataset for a 3-day period directly preceded the red team's efforts.  Similarly, for each of the four models we limited our test data to the same host used to train the model, and used records from the full three days of the red team's work, during which hosts 201 and 203 were attacked at some point, while 202 and 204 never appear to have been targeted by the red team.  This collection of pre-processing steps left us with a training set of roughly 15 million records per host, comprised of 27 fields, and a vocabulary space of roughly 6000 terms per host.  Our test set (similarly pre-processed) was roughly 9 million records per host, with roughly 37,000 and 19,000 of those being labeled as malicious on hosts 201 and 203, respectively.

\section{End-to-End Record Vectorization and Anomaly Detection Model}

\indent\indent To reiterate, the goal of this work is to build an anomaly detection scheme capable of discovering \emph{novel} cyberthreat activity, that would otherwise be missed by traditional IDS. Our NLP-inspired approach to anomaly detection requires us to embed our pre-processed records (27-dimensional vectors of vocabulary indices, with each dimension drawn from a relevant field in the ECAR records) into a semantically-relevant vector space, which is the input to our autoencoder-based anomaly detector at inference.  To that end, we sought to first vectorize the elements of our vocabulary, which would then allow us to construct relevant vector representations of the full documents composed of those elements.

\subsection{Model Architecture and Loss}

\indent\indent Instead of relying on embeddings trained on an unsupervised word2vec task, our architecture trains the word vectors inline with our autoencoder, allowing the relevant downstream task to exert pressure on the embeddings instead, producing embeddings specifically-tuned for use in the anomaly detection task.  Our ultimate model architecture passes all words (one-hot encodings, here represented by sparse indices $y_i$) in the record through a common embedder, and generates a record representation ($\mathbf{V}$) by concatenating those word vectors.  These record vectors are then passed through a single-layer autoencoder (our downstream task), which attempts to compress and decompress our record vector, creating an approximation of our encoded input ($\mathbf{\hat{V}}$).  Loss on this autoencoding task is calculated using the square of the $L_2$ norm of the difference between our input and reconstructed vectors.  

	A dense extractor layer is then applied to the autoencoder's output, which attempts to predict the original input word-by-word as probability distributions ($\hat{y}_{i}$) over the training vocabulary, with our loss being given as the total cross entropy across each term in the record.  Additionally, we found that adding a loss term of the square of the $L_2$ norm of the difference between the autoencoder's input and output helped to stabilize the training process and ensure consistent internal representations of the records.

	To ensure that both losses operated at a similar scale, a scaling parameter $\alpha$ (typically on the order of $\alpha = 5$) was used to weight the output of the internal autoencoder's reconstruction loss.  Additionally, the cross entropy loss of the outer encoder was weighted by the log of the prevalences of the correct words in the training corpus (we refer to the log of the prevalence of a given term as $w_{label}$).  Given that the corpus is very sparse, with a majority of the terms being the NULL token, this weighting encourages the word embedding to avoid overvaluing such common terms.  The overall loss for a given record is summarized as follows:

\begin{center}

$Loss = (\sum_{i=1}^{27} w_{y_{i}}*CEL(\hat{y}_{i}, y_{i})) +  \alpha * \left\| V- \widehat{V}\right\|^{2}_{2}$

\end{center}

Where the CEL between $y_i$ and $\hat{y}_i$ is calculated as the cross entropy loss between the one-hot encoding associated with the sparse index $y_i$ and the predicted distribution $\hat{y}_i$.

\begin{figure}[t!]

\includegraphics[width=\columnwidth, keepaspectratio]{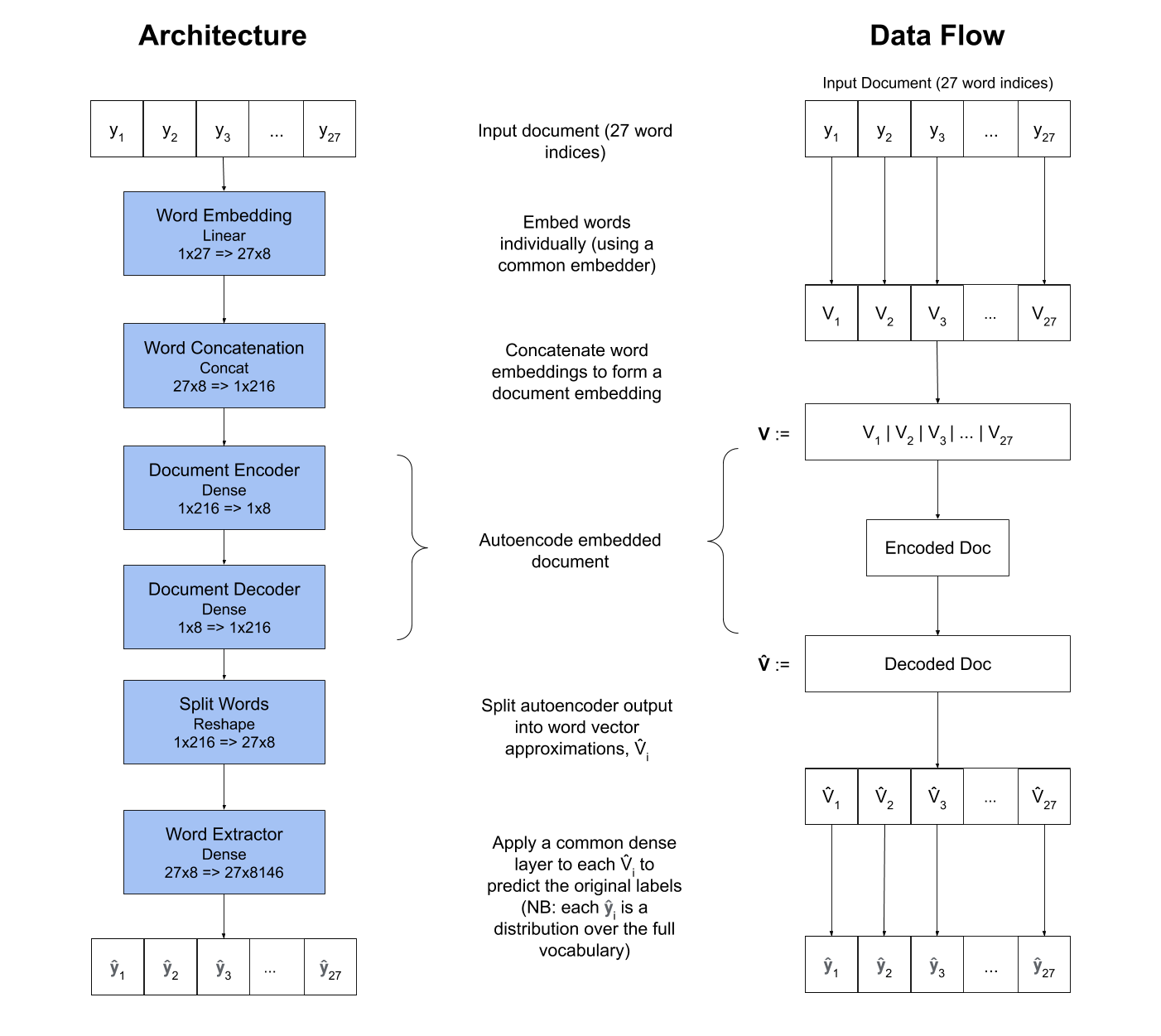}

\caption{Model architecture, with general data flow}
\label{fig:architecture}

\end{figure}

\subsection{Model Training}

\indent\indent Our model was trained on a single AWS EC2 g3xlarge instance, using Keras/Tensorflow 2.4.1.  A batch size of 32 was found to be optimal -- larger batches were tested, but the training process generally failed to produce workable results.

	Batches were generated by randomly sampling from the training corpus, with each record being weighted by the log of the prevalence of its rarest term.  This encourages the training process to sample records with rarer terms to train on, ensuring that all terms in our vocabulary are given meaningful embeddings, regardless of prevalence.

	Using the Adam optimizer, training was allowed to proceed for at most 200,000 batches, though 50,000 batches was found to produce workable results.  Training required roughly an hour using this architecture and setup, but loss stabilized relatively early (within the first 30,000 batches).

\section{Testing, Evaluation, and Results}

\indent\indent This area of research is unfortunately lacking in baseline techniques and measurements for performance (a problem that is especially exaggerated by our use of a new data set with few previous related works).  We therefore seek to provide a general baseline for the efficacy of our method, and to contextualize our results in the light of potential practical applications of our work.  Our testing methodology is inspired by the ideal end use case for our prototype -- namely, we foresee these types of utilities being used to alert administrators at times when network and host logs are exhibiting anomalous behavior, and to provide prioritized lists of anomalies for investigation/action in a SOC/administrative setting. Because of these use cases, overall accuracy is not a useful measure for our performance, as we have a large class imbalance (less than 1 percent of our test data is labeled as anomalous), and as was previously stated, our goal is not necessarily to correctly classify all of our malicious traffic as anomalous, as some of that traffic may be only incidentally labeled as malicious due to its association with a redteam process.  Instead, we choose to provide two main results that address the above use cases: a qualitative view of changes in the network's cross-entropy loss over time, and measurements of precision using various error levels as classification thresholds.  We feel that these results best promote our goal of allowing administrative users to know when an attack is taking place, while also providing a manageable number of true positives, with a high degree of certainty (i.e. high precision).

\begin{figure}[h!]
	\begin{subfigure}{\linewidth}
		\includegraphics[width=\columnwidth, keepaspectratio]{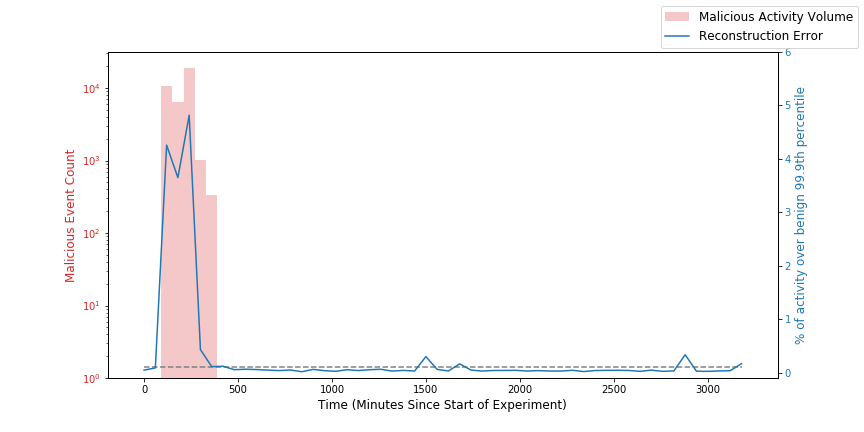}
		\caption{Host 0201}
	\end{subfigure}\hfill
	\begin{subfigure}{\linewidth}
		\includegraphics[width=\columnwidth, keepaspectratio]{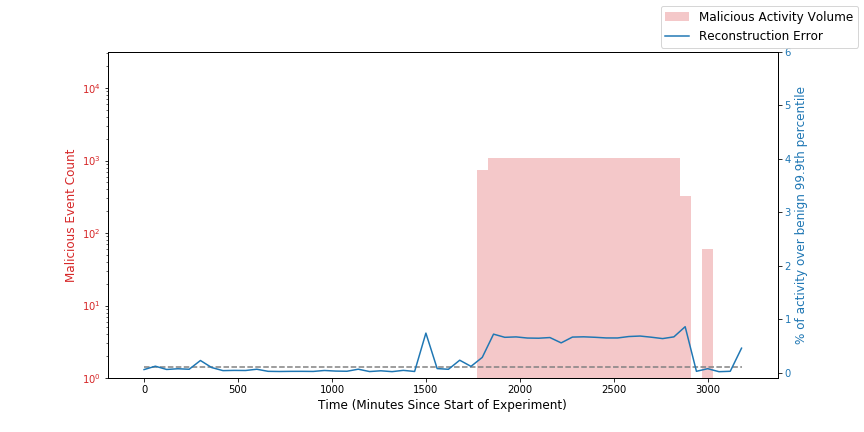}
		\caption{Host 0203}
	\end{subfigure}\hfill
	\caption{A comparison of the amount of red team activity in a given 60-minute bucket, and the amount of activity over the $99.9^{th}$ percentile of the error distribution for the training data on the associated host.  The dashed gray line is set at the average expected level of the graph - 0.1 percent.  This figure includes only the attacked hosts.}
	\label{fig:temporal_malicious}
\end{figure}

\begin{figure}[h!]
	\begin{subfigure}{\linewidth}
		\includegraphics[width=\columnwidth, keepaspectratio]{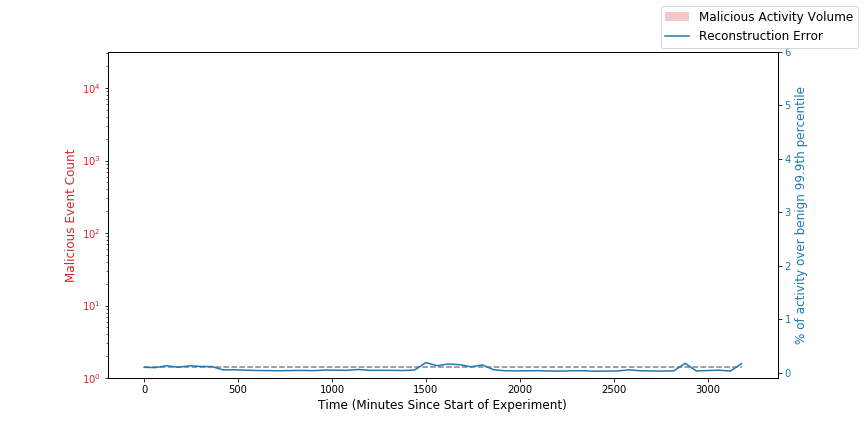}
		\caption{Host 0202}
	\end{subfigure}\hfill
	\begin{subfigure}{\linewidth}
		\includegraphics[width=\columnwidth, keepaspectratio]{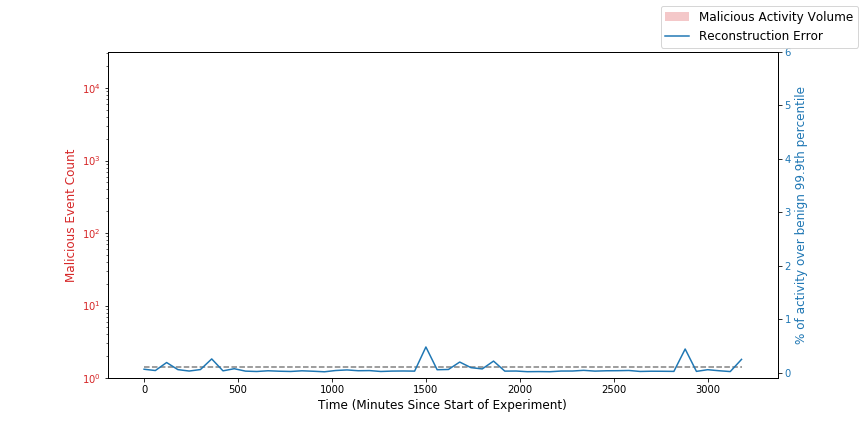}
		\caption{Host 0204}
	\end{subfigure}\hfill
	\caption{A comparison of the amount of red team activity in a given 60-minute bucket, and the amount of activity over the $99.9^{th}$ percentile of the error distribution for the training data on the associated host.  The dashed gray line is set at the average expected level of the graph - 0.1 percent.  This figure includes only the unaffected hosts, so no malicious activity is expected.}
	\label{fig:temporal_benign}
\end{figure}

The first of these use cases (our temporal alerting system) is seen in figures \ref{fig:temporal_malicious} and \ref{fig:temporal_benign}. In a SOC, an administrator may find it useful to be able to determine when a given host's activity has changed from benign to anomalous, as this may indicate the onset of an attack.  To achieve this, we gather our test data into 60-minute buckets, calculate the percentage of records in each bucket over the $99.9^{th}$ percentile of the associated training data's error distribution, and display the resulting time series in figure.  To show the correlation between this bucketed reconstruction error metric and the actual red team activity on the box, we additionally plot a histogram of the amount of red team activity in each of the sixty-minute buckets.  Consider the figure for host 201: note that the reconstruction error is relatively stable and low when there is no red team activity on the box, indicating that the embeddings and autoencoder are able to generalize to unseen new data from the same host.  Once the red team begins working, we see that the percentage of high-loss records clearly spikes upwards, indicating some degree of anomalous activity on the host.  Once the red team pivots to a new host, we see that the loss returns to its previous normal levels.  In an operationalized tool, this would enable alerting of network defenders, prompting further investigation into the causes of the spike.

To remark on one item of note: the spikes in our metrics at t=1500 and t=2880 are correlated with large (on the order of a 10-fold increase) spikes in the volume of WMI (Windows Management Instrumentation) activity on each host.  This is anomalous in and of itself, and it is true that the red team made broad use of WMI between these two spikes to pivot throughout the network, but it is unclear if this activity can be explicitly tied to the red team actors due to the coarse-grain nature of the red team's ground truth documentation.  As such, we have elected to treat these spikes as non-malicious anomalies.

\begin{figure}[h!]
	\begin{subfigure}{\linewidth}
		\includegraphics[width=\columnwidth, keepaspectratio]{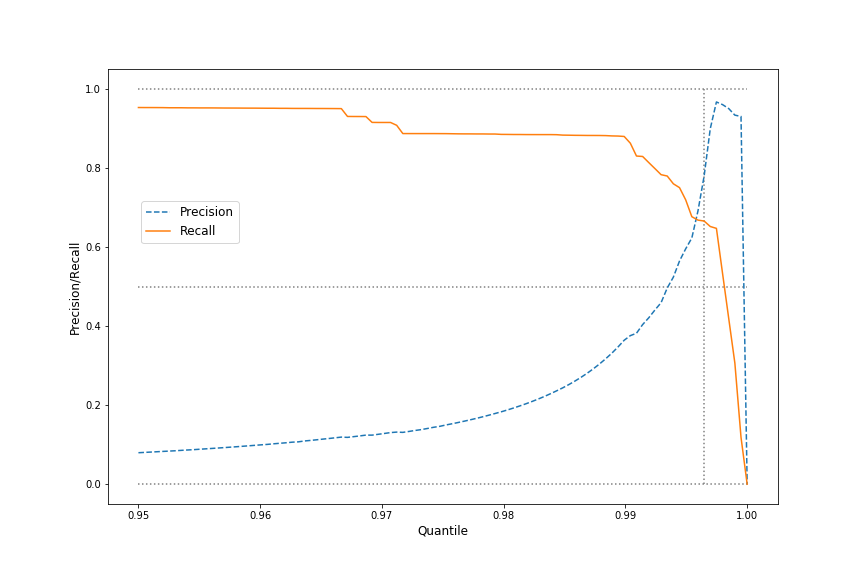}
		\caption{Host 0201}
	\end{subfigure}
	\begin{subfigure}{\linewidth}
		\includegraphics[width=\columnwidth, keepaspectratio]{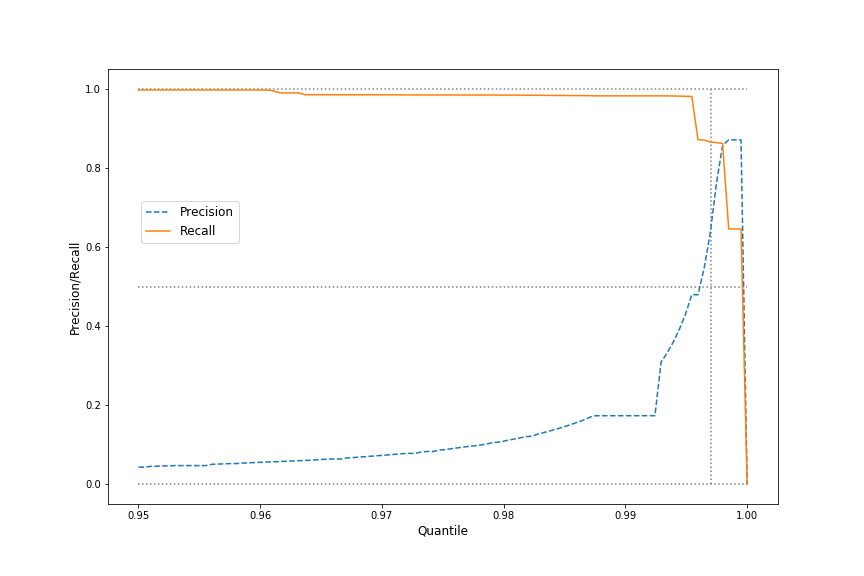}
		\caption{Host 0203}
	\end{subfigure}
	\caption{Test precision and recall by quantile threshold.  The vertical dotted line marks the $99.9^{th}$ percentile of the error distribution of the associated host's benign training data.  Note that 80 and 66 percent precision with 66 and 86 percent recall can be achieved at these thresholds for hosts 201 and 203, respectively.}
	\label{fig:thresholds}
\end{figure}

Our second metric provides a more quantitative view of our results by indicating the precision of classifying the top q percent (in terms of cross entropy loss) of the data set as anomalous.  This evaluation is performed as follows:

\begin{enumerate}
	\item Pre-process and vectorize test data by embedding the associated terms and concatenating those embeddings
	\item Pass those test vectors through our trained network, and calculate the cross-entropy loss for each record
	\item For each quantile, q, between the 95th and 100th percentile of our test loss distribution, classify every record with a loss above q as being anomalous.
	\item For each threshold, q, calculate the precision of our classification.  Recall is additionally calculated for completeness.

\end{enumerate} 

	Figure \ref{fig:thresholds} shows the classification precision by quantile threshold for each of the attacked hosts-- by setting a threshold at the 99.9th percentile of our training reconstruction error distributions, we can see that our classification system provides at least 66 percent precision; the host that saw the most involved attack (that is, host 201 - host 203 simply exhibited beaconing behavior), we saw 80 percent precision.  In a realistic tool built on such work, this would ensure that examining a ranked list of anomalies discovered by the system would quickly yield obvious true anomalies.  For additional context, this classification method would have had a 0.082 percent and 0.096 percent false discovery rate on the unattacked hosts 202 and 204, respectively, which is in-line with the 99.9 percentile threshold, indicating that the error distributions are consistent across the training and test sets, and that ordinary activity should not be overly-given to produce false positives.

\begin{figure}[b!]
	\begin{subfigure}{\linewidth}
		\includegraphics[width=\columnwidth, keepaspectratio]{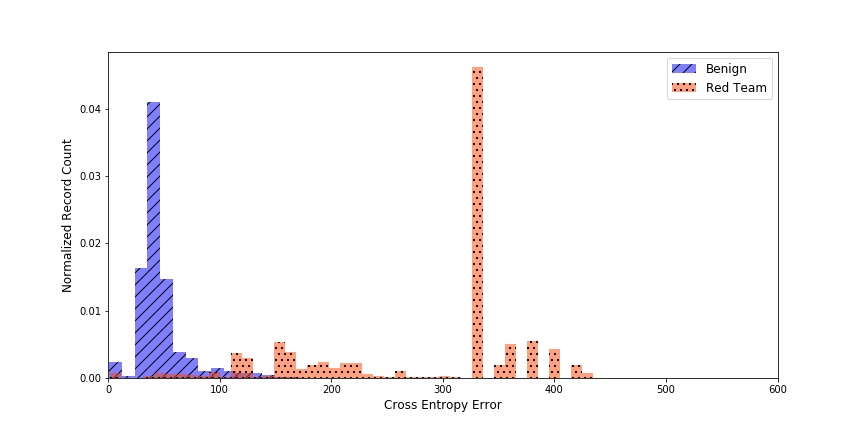}
		\caption{Host 201}
	\end{subfigure}
	\begin{subfigure}{\linewidth}
		\includegraphics[width=\columnwidth, keepaspectratio]{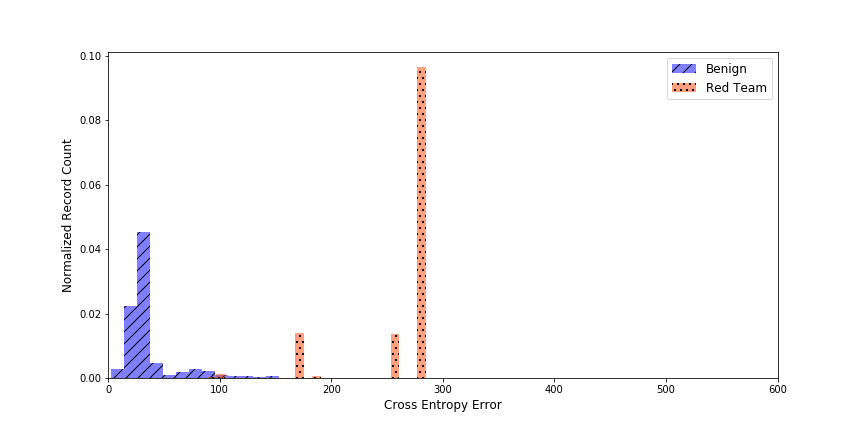}
		\caption{Host 203}
	\end{subfigure}

\caption{A histogram of reconstruction errors on the attacked hosts, segregated by class: benign and malicious -- note that the two populations have been normalized to keep them on the same scale in each case.}
\label{fig:histogram_malicious}

\end{figure}

\begin{figure}[b!]
	\begin{subfigure}{\linewidth}
		\includegraphics[width=\columnwidth, keepaspectratio]{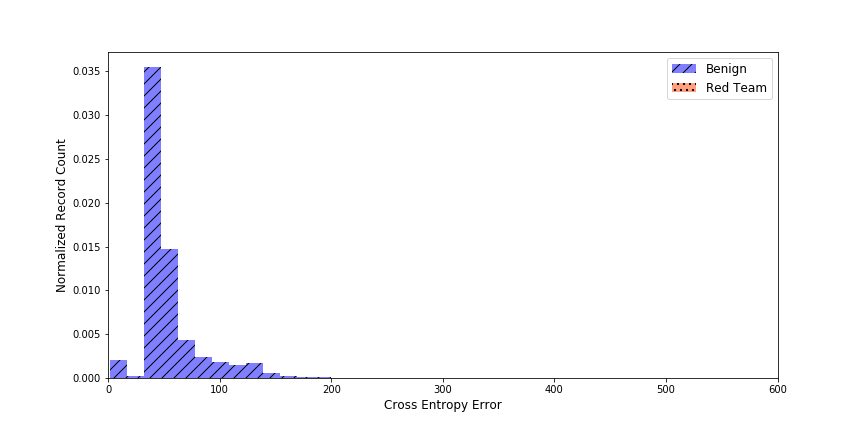}
		\caption{Host 202}
	\end{subfigure}
	\begin{subfigure}{\linewidth}
		\includegraphics[width=\columnwidth, keepaspectratio]{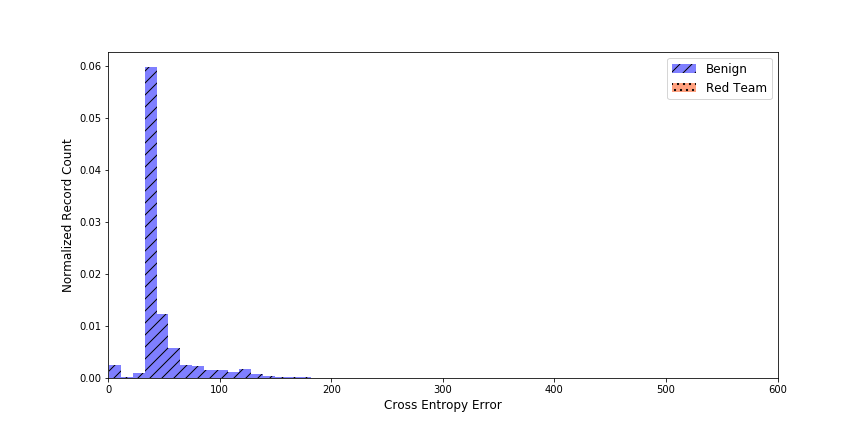}
		\caption{Host 204}
	\end{subfigure}

\caption{A histogram of reconstruction errors on the benign hosts -- note that the distribution has been normalized.}
\label{fig:histogram_benign}

\end{figure}

	To provide a better view of the range and distribution of the reconstruction errors, we additionally provide a histogram of the errors (figures \ref{fig:histogram_malicious} and \ref{fig:histogram_benign}), segregated by class.  This shows a clear separation between a large portion of the malicious activity, and the majority of the remaining activity.

\begin{figure}[t!]

\includegraphics[width=\columnwidth, keepaspectratio]{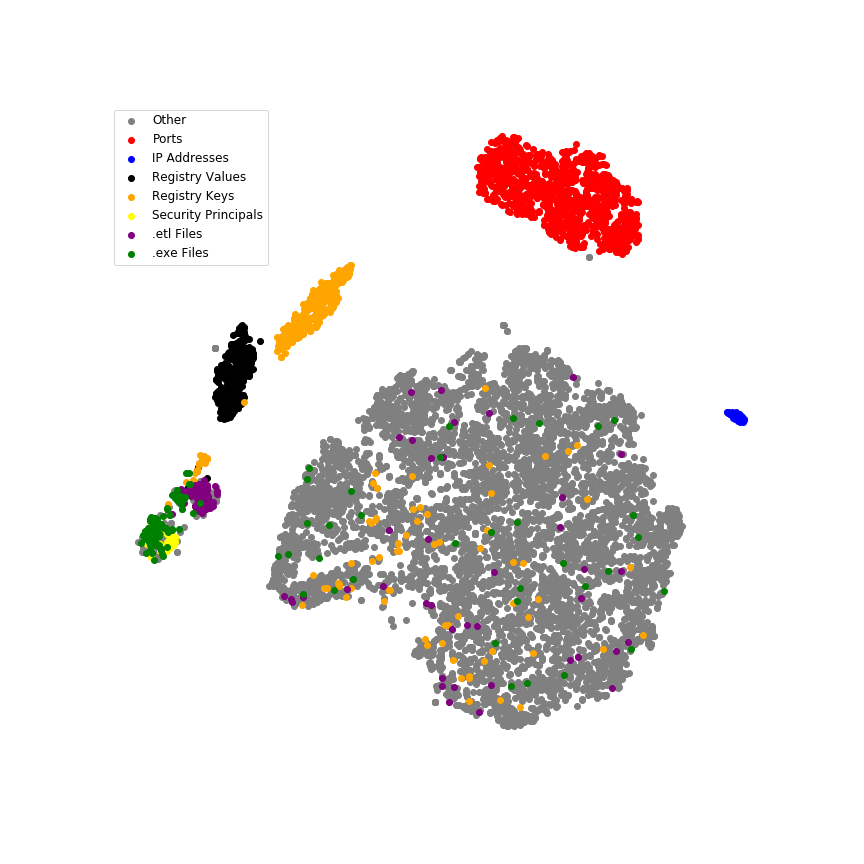}

\caption{A TSNE visualization of the word embeddings learned by the model.  Labels are provided for relevant categories of words.}
\label{fig:embeddings}

\end{figure}

	Finally, to provide a more qualitative context for our results, we additionally provide a visualization of the word vector embeddings learned by the system for host 201 (figure \ref{fig:embeddings}).  This is done by extracting the embeddings from the model, reducing their dimensionality using T-SNE, and labeling the emergent structure discovered within the visualization.

\begin{figure}[t!]

\includegraphics[width=\columnwidth, keepaspectratio]{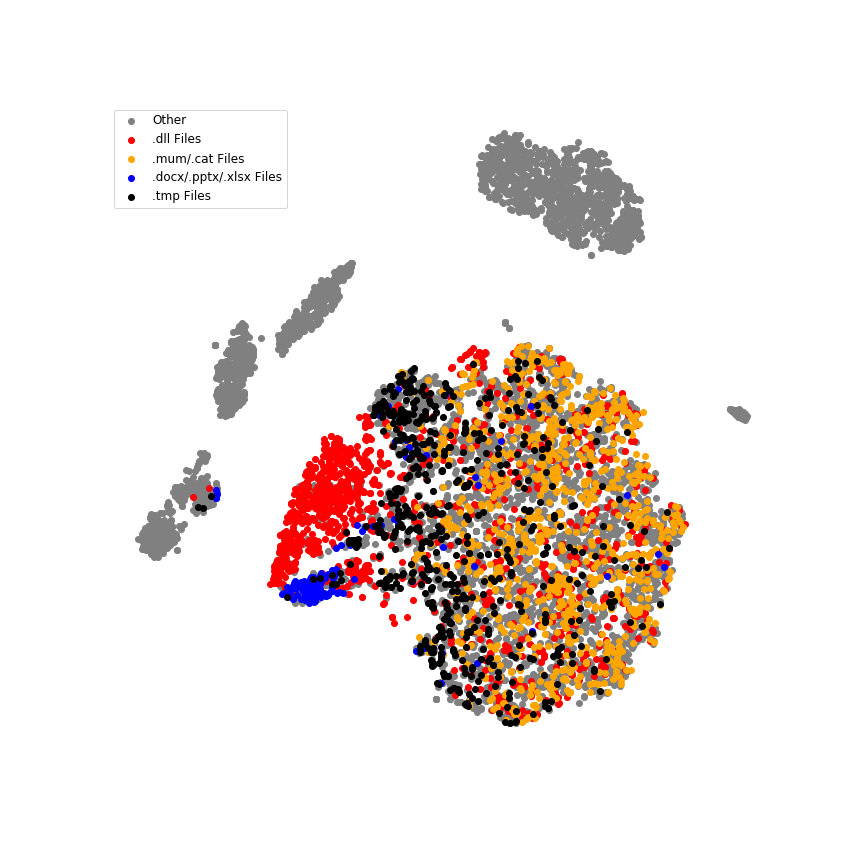}

\caption{A breakdown of the large cluster in our TSNE visualization -- note the clustering of of common types of files, including Microsoft Office suite files, DLLs, .tmp files created by the system, and .mum/.cat files associated with the Microsoft update process.}
\label{fig:detailed_embeddings}

\end{figure}

	Figure \ref{fig:embeddings} shows that there is clear role-based clustering of terms, both network-related (e.g. IP addresses and ports) and host-related (executables, logs, registry information, etc.).  The large gray cluster in \ref{fig:embeddings} largely contains embeddings of terms that refer to files and resources being manipulated by the system in vairous ways.  For instance, DLL modules loaded by the system, Microsoft Office files modified by the Office suite's software, and .tmp files can be found in the cluster.  These file types, among others, combine into subclusters, which we show in figure \ref{fig:detailed_embeddings}.  Because files can be manipulated in many ways (e.g. DLL files can be loaded as modules, or they can be moved to a new directory), the clustering is less clear than the independent clusters noted in \ref{fig:embeddings}, but the role-based groupings are still clear.  Our training process's generation of these semantically-relevant embeddings shows that our autoencoding process is learning relevant patterns for the compression and reconstruction of our records, meaning that the measures discussed above (precision, etc.) are relevant to our end goal.

\section{Future Work}

\indent\indent This work has shown that there is some promise in this general end-to-end approach to anomaly detection, but further investigation of several key issues is still needed.

\begin{enumerate}

	\item Vocabulary improvement: Generally, we want to remove as much ad-hoc feature engineering as possible from our process to ensure that we do not require or depend on strict assumptions or models.  However, our current approach requires a higher-than-necessary amount of vocabulary engineering (e.g. rare term thresholding and combination, term discretization, file path simplification etc.).  Working to remove these ad-hoc aspects of our feature engineering regimen is a high priority.
	\item Network architecture improvement: our current architecture is suited for its role as a proof-of-concept, but it is hardly state of the art -- future work should work to improve our design by improving the following:
	\begin{itemize}
		\item Increasing the depth/capacity of our autoencoder as appropriate to capture broader structure.
		\item Encouraging the development of structure in the latent space of our autoencoder (e.g. clustering, topological structure, etc.) to strengthen the meanfulness of the latent space (and therefore, the meaninfulness of the conclusions drawn on the basis of that space).
		\item Exploring alternate methods for record embedding beyond simple concatenation, such as doc2vec-style record embeddings.  Additionally, aggregating records across time by process ID or user should be explored as a method for contextualized, behavior-based analysis.  Appropriately, the application of recurrent, convolutional, and attentional networks should be explored in this space.
	\end{itemize}
	\item Online training: Our current approach does not account for the streaming nature of our data source, and so could not truly be utilized as an effective operational tool.  Future work should explore methods for timely training to prevent model staleness, as well as methods for accounting for the appearance of new vocabulary and phenomena over time.
	\item Model explainability: Our work does not focus on explainability, but given its intended nature as a portion of a human/machine teaming utility, explainability should be developed with some urgency.  Analysis of the contributing components of the autoencoder's reconstruction error could provide an intuition as to why a given record is considered to be anomalous.  Alternately, examining the components of the end-to-end cross entropy loss in the model could yield similar outcomes.

\end{enumerate}

\section{Conclusion}

\indent\indent In this work, we demonstrated an end-to-end autoencoder model that leverages techniques borrowed from natural language processing to detect anomalies in a broad collection of network and host log data.  Red team activity in our test set was clearly detected, with the raw reconstruction error distributions and temporal summarizations providing obvious indications of anomalous activity.  Our classification method exhibits the desired degree of precision, showing the potential for practical implementations based on this work.  Additionally, the embeddings trained through this end-to-end process show clear signs of semantically-relevant structure, indicating that our architecture and training process is capable of learning coherent descriptions of a given host's activity.

\appendix
\section{List of ECAR Keys Used in Experiments}\label{app:ECAR}
\begin{enumerate}
	\item action
	\item command\textunderscore line
	\item dest\textunderscore ip
	\item dest\textunderscore port
	\item direction
	\item file\textunderscore path
	\item image\textunderscore path
	\item info\textunderscore class
	\item key
	\item l4protocol
	\item logon\textunderscore id
	\item module\textunderscore path
	\item new\textunderscore path
	\item object
	\item parent\textunderscore image\textunderscore path
	\item path
	\item principal
	\item requesting\textunderscore domain
	\item requesting\textunderscore logon\textunderscore id
	\item requesting\textunderscore user
	\item sid
	\item task\textunderscore name
	\item type
	\item user
	\item user\textunderscore name
	\item value
	\item duration (calculated from flow start and end times)
\end{enumerate}

\bibliographystyle{named}
\bibliography{refs}

\end{document}